\documentclass[10pt,twocolumn,letterpaper]{article}

\usepackage[pagenumbers]{cvpr} 
\usepackage{url}

\usepackage{tabularx}
\usepackage{multirow}
\usepackage{siunitx}
\usepackage[normalem]{ulem}

\usepackage[table]{xcolor}
\usepackage{xcolor}

\usepackage{pifont}
\newcommand{\cxmark}{\ding{51}}

\newlength{\orgtabcolsep}
\usepackage{arydshln}

\newcommand{\x}[0]{\boldsymbol{x}}
\newcommand{\mean}[0]{\boldsymbol{\mu}}
\newcommand{\var}[0]{\boldsymbol{\Sigma}}

\newcommand{\1}[0]{\cellcolor[HTML]{AAFFAA}\bfseries}
\newcommand{\2}[0]{\cellcolor[HTML]{EEFF55}\hspace{-0.3em}}
\newcommand{\3}[0]{\cellcolor[HTML]{FFCC55}\hspace{-0.3em}}

\makeatletter
\DeclareRobustCommand\onedot{\futurelet\@let@token\@onedot}
\def\@onedot{\ifx\@let@token.\else.\null\fi\xspace}

\def\eg{\emph{e.g}\onedot}

\makeatother

\renewcommand{\paragraph}[1]{\noindent\textbf{#1.}}

\definecolor{cvprblue}{rgb}{0.21,0.49,0.74}
\usepackage[pagebackref,breaklinks,colorlinks,citecolor=cvprblue]{hyperref}

\def\confName{3DV\xspace}
\def\confYear{2025\xspace}

\title{E-3DGS: Event-Based Novel View Rendering of Large-Scale Scenes\\Using 3D Gaussian Splatting} 

\author{Sohaib Zahid$^{1,2}$ \quad Viktor Rudnev$^{1,2}$ \quad Eddy Ilg$^1$ \quad Vladislav Golyanik$^2$ \\
\vspace{5pt}
$^1$Saarland University \quad $^2$MPI for Informatics, SIC}

\begin{document}
\maketitle

\begin{abstract} 
Novel view synthesis techniques predominantly utilize RGB cameras, inheriting their limitations such as the need for sufficient lighting, susceptibility to motion blur, and restricted dynamic range. In contrast, event cameras are significantly more resilient to these limitations but have been less explored in this domain, particularly in large-scale settings. Current methodologies primarily focus on front-facing or object-oriented (360-degree view) scenarios. For the first time, we introduce 3D Gaussians for event-based novel view synthesis. Our method reconstructs large and unbounded scenes with high visual quality. We contribute the first real and synthetic event datasets tailored for this setting. Our method demonstrates superior novel view synthesis and consistently outperforms the baseline \mbox{EventNeRF} by a margin of $11{-}25\%$ in PSNR (dB) while being orders of magnitude faster in reconstruction and rendering. 
\end{abstract} 

\section{Introduction}\label{sec:Intro}

Novel view synthesis offers a fundamental approach to visualizing complex scenes by generating new perspectives from existing imagery. 
This has many potential applications, including virtual reality, movie production and architectural visualization \cite{Tewari2022NeuRendSTAR}. 
An emerging alternative to the common RGB sensors are event cameras, which are  
 bio-inspired visual sensors recording events, i.e.~asynchronous per-pixel signals of changes in brightness or color intensity. 

Event streams have very high temporal resolution and are inherently sparse, as they only happen when changes in the scene are observed. 
Due to their working principle, event cameras bring several advantages, especially in challenging cases: they excel at handling high-speed motions 
and have a substantially higher dynamic range of the supported signal measurements than conventional RGB cameras. 
Moreover, they have lower power consumption and require varied storage volumes for captured data that are often smaller than those required for synchronous RGB cameras \cite{Millerdurai_3DV2024, Gallego2022}. 

The ability to handle high-speed motions is crucial in static scenes as well,  particularly with handheld moving cameras, as it helps avoid the common problem of motion blur. It is, therefore, not surprising that event-based novel view synthesis has gained attention, although color values are not directly observed.
Notably, because of the substantial difference between the formats, RGB- and event-based approaches require fundamentally different design choices. %

The first solutions to event-based novel view synthesis introduced in the literature demonstrate promising results \cite{eventnerf, enerf} and outperform non-event-based alternatives for novel view synthesis in many challenging scenarios. 
Among them, EventNeRF \cite{eventnerf} enables novel-view synthesis in the RGB space by assuming events associated with three color channels as inputs. 
Due to its NeRF-based architecture \cite{nerf}, it can handle single objects with complete observations from roughly equal distances to the camera. 
It furthermore has limitations in training and rendering speed: 
the MLP used to represent the scene requires long training time and can only handle very limited scene extents or otherwise rendering quality will deteriorate. 
Hence, the quality of synthesized novel views will degrade for larger scenes. %

We present Event-3DGS (E-3DGS), i.e.,~a new method for novel-view synthesis from event streams using 3D Gaussians~\cite{3dgs} 
demonstrating fast reconstruction and rendering as well as handling of unbounded scenes. 
The technical contributions of this paper are as follows: 
\begin{itemize}
\item With E-3DGS, we introduce the first approach for novel view synthesis from a color event camera that combines 3D Gaussians with event-based supervision. 
\item We present frustum-based initialization, adaptive event windows, isotropic 3D Gaussian regularization and 3D camera pose refinement, and demonstrate that high-quality results can be obtained. %

\item Finally, we introduce new synthetic and real event datasets for large scenes to the community to study novel view synthesis in this new problem setting. 
\end{itemize}
Our experiments demonstrate systematically superior results compared to EventNeRF \cite{eventnerf} and other baselines. 
The source code and dataset of E-3DGS are released\footnote{\url{https://4dqv.mpi-inf.mpg.de/E3DGS/}}.

\section{Related Work}\label{sec:Related_Work}

\subsection{Novel View Synthesis from RGB Inputs}\label{ssec:RW_RGB}
Novel view synthesis of rigid scenes is predominantly handled assuming RGB inputs. A widely used approach to this problem is to learn coordinate-based neural scene representations allowing rendering novel views at test time. 
Earlier works such as Neural Radiance Fields (NeRF) and its direct follow-ups \cite{nerf, Tewari2022NeuRendSTAR} used implicit neural representations in combination with volume rendering.
They are based on expensive-to-optimize Multi-Layer Perceptrons (MLPs) and are slow at training and evaluation while requiring a relatively low amount of storage space once they are trained.
Their stochastic ray sampling requires many samples to obtain an accurate scene approximation, and shooting rays through empty space constitutes unnecessary overhead. Most of these approaches focus on single objects or bounded scenes. 
Recent techniques accelerate neural MLP-based representations or ray sampling \cite{Reiser2021ICCV, instantngp} or avoid MLPs \cite{Sun2022DirectVG, YuFridovichKeil2022, zipnerf} by using voxel grids. 
Some techniques~\cite{neurbf} support unbounded scenes by employing radial basis functions, thereby overcoming the limitations of voxel-grid-based methods. 
Several ray tracing-based methods support large-scale scenes and uncontrolled camera trajectories thanks to progressive NeRF optimization \cite{xiangli2022bungeenerf, meuleman2023localrf}.
Instant-NGPs \cite{instantngp} are neural feature volumes with a hash grid that can be learned and evaluated quickly at test time. They can also handle multi-scale training scenarios efficiently.

A promising recent development is the shift from ray tracing to rasterization, marked by the introduction of 3D Gaussian Splatting (3DGS) \cite{3dgs}.
This approach presents an alternative paradigm for 3D reconstruction and novel view synthesis using differentiable rasterization with 3D Gaussians as geometric primitives.
Since GPU technology and algorithmic research have evolved over several decades to provide high performance for rasterization applications,
3DGS trains substantially quicker and provides much higher rendering throughput than NeRF. Moreover, since it explicitly represents the geometry, it can scale easily as the scene size increases with no special handling required for unbounded scenes. Our approach adopts the 3D Gaussian representation and presents its application to the supervision from event streams. It inherits thereby the advantages of event streams and 3DGS for view synthesis. 

\subsection{Novel View Synthesis from Event Streams}
Event-aided sparse odometry and simultaneous localization and mapping approaches are distantly related to our setting, as they do not allow photo-realistic and dense rendering of novel views \cite{Kim2016, Rebecq2017EVOAG, Hidalgo-Carrio_2022_CVPR, Klenk2024DeepEV}.

As previously discussed, event cameras represent an alternative to RGB sensors for dense novel view synthesis, and some initial work was done on learning 3D scene representations from event streams only. EventNeRF \cite{eventnerf} is a seminal framework for training MLP-based implicit 3D representations (see Sec.~\ref{ssec:RW_RGB}) using frames of accumulated color events.
While it demonstrates impressive results, it is restricted to camera trajectories with uniform motion and the assumption that the background is a constant color (triggering no events).
E-NeRF \cite{enerf} is another work that resembles the training methodology of EventNeRF for single-channel (intensity) event cameras and allows training a colored 3D representation from a combination of blurry RGB images and grayscale events. Robust E-NeRF by Low and Lee \cite{robust_enerf} is a model aiming to reduce the issues caused by uncontrolled camera motion. They introduce the refractory period to the event generation model, i.e.~the time during which a pixel is inactive after an event firing. Supervision happens on the level of individual events, and they reformulate the event loss to handle intra-pixel variances of the contrast threshold optimized during training. All these methods adopt ray tracing and can be primarily applied on 360\textdegree~object-centric  datasets or front-facing trajectories.

Our approach differs from previous event-based methods in that it demonstrates that rasterization can be efficiently combined with event-based supervision instead of ray tracing.
The main design choices of our method are tailored to 3D Gaussians. As a result, our method inherits the primary advantages of 3DGS \cite{3dgs}, such as fast training and inference. Similar to EventNeRF \cite{eventnerf}, our method supports color. However, in contrast, it is not limited to single objects and can handle large-scale scenes.

\begin{figure*}[t]
\centering
\includegraphics[width=1.0\linewidth]{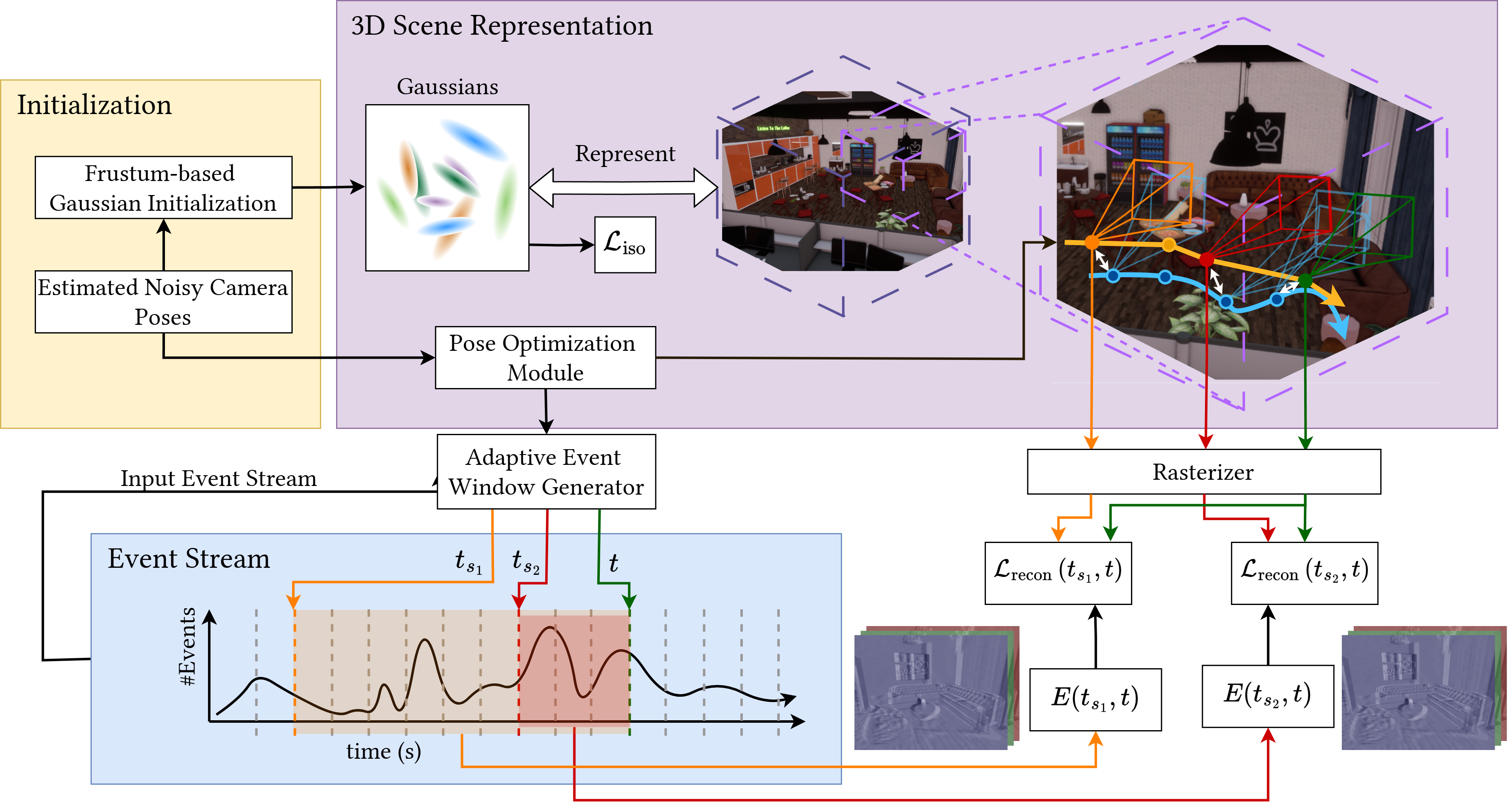}
\caption{
\textbf{Overview of our E-3DGS Method.}
We use 3D Gaussians~\cite{3dgs} as the scene representation and assume that initial noisy camera poses are available. We randomly initialize the scene with our frustum-based initialization (Sec.~\ref{sec:frustum_init}) and then optimize the Gaussians and the camera poses jointly (Sec.~\ref{sec:pose_refinement}). To obtain a high-quality reconstruction of both, low-frequency structure and high-frequency detail, we propose a strategy using a large event window from $t_{s_1}$ to $t$ and a small one from $t_{s_2}$ to $t$ (Sec.~\ref{subsec:adaptive_window}). We then define the loss $\mathcal{L}_\mathrm{recon}$ (Sec.~\ref{ssec:Optimization}) between renderings from our model at the current time $t$ (indicated green) and previous times $t_{s_1}$ (indicated orange) and $t_{s_2}$ (indicated red), and the accumulated incoming events $E(t_{s_1},t)$ and $E(t_{s_2},t)$. We regularize the 3D Gaussians with the loss $\mathcal{L}_\mathrm{iso}$ (Sec.~\ref{ssec:IsotropicReg}). 
}
\label{fig:methodology}
\end{figure*}

\section{Preliminaries}\label{sec:Preliminaries} 

\subsection{3D Gaussians}
3D Gaussian Splatting~\cite{3dgs} is a high-quality and efficient scene representation. 
The Gaussians are defined by a 3D covariance matrix \(\var_i\) centered around a point \(\mean_i\): 
\begin{equation}
   G_i(\x) = \exp\left(-\frac{1}{2}(\x - \mean_i)^T \var_i^{-1} (\x - \mean_i)\right) \mathrm{\,,}
   \label{eq:gaussian_density}
\end{equation}
and their overlay models the 
geometry at scene location~$\x$.
Each Gaussian is additionally associated with an opacity~$o_i$ and spherical harmonics that model view-dependent color.
For rendering purposes, the means~$\mean_i$ and covariance matrices~$\var_i$ are transformed into image coordinates. The projected matrix~$\var'_i$ can be obtained by applying the viewing transformation~$\boldsymbol{W}$ and the Jacobian~$\boldsymbol{J}$ of the affine approximation of the projective transformation:
\begin{equation}
   \var'_i = \boldsymbol{J} \boldsymbol{W} \var_i \boldsymbol{W}^T \boldsymbol{J}^T \mathrm{\,.}
   \label{eq:gaussian_project}
\end{equation}
The third row and column of~$\var'_i$ are dropped to obtain a 2D matrix.
Using Equation~\ref{eq:gaussian_density}, one can then evaluate the different Gaussians~$i$ that overlap with an image pixel~$x$ and obtain alpha values as \(\alpha_{i,x} = o_i G'_i(\x)  \). 
The Gaussians are then sorted according to their depth, and alpha blending for every pixel  is performed by combining the view-dependent colors $c_i$ using the following equation:
\begin{equation}
   C_x = \sum_{i=1}^{N} T_{i,x} \alpha_{i,x} c_i \mathrm{\,,}
\end{equation}
where \(T_{i,x} = \prod_{k=1}^{i-1} (1 - \alpha_{k, x})\) represents the transmittance. 

\subsection{Event Formation Model}

Event cameras generate a continuous stream of events denoted as $ e = (\x, p, \tau) $, where $\x$ are the pixel coordinates at which an event is triggered at time $\tau$, and $p \in \{-1, +1\}$ signifies the polarity of the event, indicating an increase or decrease in the logarithmic intensity by 
the predefined contrast threshold \( \Delta \). 
Thus, the relationship between the triggered event and the logarithmic image intensity reads: 
\begin{equation}
    L_{\x}(\tau) - L_{\x}(\tau^\mathrm{prev}) = p\Delta, 
    \label{eq:egm}
\end{equation}
where \( \tau^\mathrm{prev} \) is the time when the previous event for the pixel was triggered. This concept can then be generalized to apply for an accumulation of events within a time interval $(\tau_1, \tau_2) | \tau_1 < \tau_2$ for a pixel location $\x$ as follows:
\begin{equation}
    L_{\x}(\tau_2) - L_{\x}(\tau_1) = \sum_{\tau_1 < \tau_t \leq \tau_2} p_t\Delta \stackrel{\mathrm{def}}{=} E_{\x}(t_1, t_2)\mathrm{\,,}
    \label{eq:egm_sum}
\end{equation}
where $t_1$, $t_2$ index the sequence of events closest to 
$\tau_1$, $\tau_2$.

\section{The E-3DGS Method}\label{sec:Method} 
Our aim is to learn a 3D representation of a static scene using only a color event stream, where each pixel observes changes in brightness corresponding to one of the red, green, or blue channels according to a Bayer pattern, with known camera intrinsics $K_t~\in~\mathbb{R}^{3 \times 3}$, and noisy initial poses~$P_t~\in~\mathbb{R}^{3 \times 4}$, at reasonably high-frequency time steps indexed by $t$. 
Following 3DGS~\cite{3dgs}, we represent our scene by anisotropic 3D Gaussians. Our methodology comprises a technique to initialize Gaussians in the absence of a Structure from Motion (SfM) point cloud, adaptive event frame supervision of 3DGS, and a pose refinement module. 
An overview of our method is provided in Fig.~\ref{fig:methodology}.

Our E-3DGS method is not restricted to scenes of a certain size and can handle unbounded environments. It does not rely on any assumptions regarding the background color, type of camera motion, or speed. Thus, it ensures robust performance across a wide range of scenarios. 

\subsection{Event Stream Supervision} 

There are two main categories of approaches to learning 3D scene representations from event streams. 
Some apply the loss to single events~\cite{robust_enerf} based on Eq.~\eqref{eq:egm}. Others use the sum of events~\( E_{\x}(t_1,t_2) \) from Eq.~\eqref{eq:egm_sum}. We choose the second approach, as rasterization in 3DGS is well suited to efficiently render entire images rather than individual pixels. 

To optimize our Gaussian scene representation using event data, we can make a logical equivalence between the observed event stream and the scene renderings. 
To do so, we replace the true logarithmic intensities~\( L_{\x} \) in Eq.~\eqref{eq:egm_sum} with the rendered logarithmic intensities~\(\hat{L}_{\x} \) from our scene, and the times $\tau$ with the camera poses $P_t$ that were used to render the scene at the respective time steps. 
Following the approach used in~\cite{eventnerf}, the log difference is then point-wise multiplied with a Bayer filter $F$ to obtain the respective color channel. We can finally calculate the error between the logarithmic change from our model and the actual change observed from the event stream, and define the following per-pixel loss: 
\begin{equation}
    \begin{split}
    &\mathcal{L}_{\x}\left(t_1, t_2\right) = \\
    &\left\| 
    F \odot (\hat{L}_{\x}(P_{t_2}) - \hat{L}_{\x}(P_{t_1})) 
    - F \odot E_{\x}\left(t_1, t_2\right)\right\|_1, 
    \end{split}
    \label{eq:L_recon_per_pixel} 
\end{equation}
where ``$\odot$'' denotes pixelwise multiplication.

\subsection{Frustum-Based Initialization}
\label{sec:frustum_init}

In the original 3DGS \cite{3dgs}, the Gaussians are initialized using a point cloud obtained from applying SfM on the input images. 
The authors also experimented with initializing the Gaussians at random locations within a cube. While this worked for them with a slight performance drop, it requires an assumption about the extent of the scene. 

Applying SfM directly to event streams is more challenging than RGB inputs \cite{Kim2016} and exploring this aspect is not the primary focus of this paper. 
In the absence of an SfM point cloud, we use the randomly initialized Gaussians and extend this approach to unbounded scenes. 
To this end, we initialize a specified number of Gaussians (on the order of \qty{d4}{}) in the frustum of each camera. 
This gives two benefits: 1) All the initialized Gaussians are within the observable area, and 2) We only need one loose assumption about the scene, which is the maximum depth $z_\mathrm{far}$.

\subsection{Adaptive Event Window}\label{subsec:adaptive_window}

Rudnev et al.~\cite{eventnerf} demonstrated in EventNeRF that using a fixed event window duration results in suboptimal reconstruction. They find that larger windows are essential for capturing low-frequency color and structure, and smaller ones are essential for optimization of finer high-frequency details. While they randomly sampled the event window duration, a drawback is that it does not consider the camera speed and event rate, thus the sampled windows may contain too many or too few events.  
As our dataset features variable camera speeds, we improve upon this by sampling the number of events rather than the window duration.  
To achieve this, for each time step we randomly sample a target number of events from within the range $[N_\mathrm{min}, N_\mathrm{max}]$. 
Given a time step~$t$, we search for a previous time step~$t_s$ such that the number of events in the event frame $E(t_s, t)$ is approximately equal to the desired number. 

When determining $N_\mathrm{max}$, we find that for values where details and low-frequency structure are optimal, 3DGS tends to get unstable and sometimes prunes away Gaussians in homogeneous areas.
While this can be mitigated by choosing a much larger $N_\mathrm{max}$, this again deteriorates the details. 
Therefore, we propose a strategy to incorporate both, small and large windows. For each $t$, we choose two earlier time steps~$t_{s_1}$ and~$t_{s_2}$. The ranges for sampling the event counts for both are empirically chosen to be $[\frac{N_\mathrm{max}}{10}, N_\mathrm{max}]$ and $[\frac{N_{max}}{300}, \frac{N_\mathrm{max}}{30}]$. We then render frames from our model at times $t$, $t_{s_1}$ and $t_{s_2}$, and use two concurrent losses for the event windows $E_{\x}\left(t_{s_1}, t\right)$ and $E_{\x}\left(t_{s_2}, t\right)$. 

\subsection{As-Isotropic-As-Possible Regularization} 
\label{ssec:IsotropicReg} 

In 3DGS, Gaussians are unconstrained in the direction perpendicular to the image plane. 
This lack of constraint can result in elongated and overfitted Gaussians. 
And while they may appear correct from the training views, they introduce significant artifacts when rendered from novel views by manifesting as floaters and distortions of object surfaces. 
We also observe that the lack of multi-view consistency and tendency to overfit destabilize the pose refinement. 

To mitigate these issues, we draw inspiration from Gaussian Splatting SLAM~\cite{3dgsslam} and SplaTAM~\cite{splatam}, and apply isotropic regularization:
\begin{equation}
    \mathcal{L}_{\text{iso}} = \frac{1}{|\mathcal{G}|} \sum_{g \in \mathcal{G}} \left\| S_g - \bar{S}_g \right\|_1
    \label{eq:L_iso} \mathrm{\,,}
\end{equation}
where~$\mathcal{G}$ is the set of Gaussians visible in the image. Eq.~\eqref{eq:L_iso} imposes a soft constraint on the Gaussians to be as isotropic as possible.
We find that it helps to improve pose refinement, minimizes floaters and enhances generalizability. 

\subsection{Pose Refinement} 
\label{sec:pose_refinement}

To obtain the most accurate results, we allow the poses to be refined during optimization
by modeling the refined pose as $P'_t = P^e_t P_t$, where  $P^e_t$ is an error correction transform. 
Instead of directly optimizing~$P^e_t$ as a~$3 \times 3$ matrix, following Hempel et al.~\cite{6d_rotation} we represent it as $[r_1\,\, r_2\,\, T]$, where $r_1$ and $r_2$ represent two rotation vectors of the rotation matrix~$R = [r_1\,\, r_2\,\, r_3]$, while $T$ is the translation.
We can then obtain the~$P^e_t$ matrix from the representation using Gram-Schmidt orthogonalization (see details in Supplement~\ref{sec:supp_pose_refinement}), hence ensuring that during optimization, our error correction transform always represents a valid transformation matrix. 
$P^e_t$ is initialized to be the identity transform. Since the loss function from Eq.~\eqref{eq:L_recon_per_pixel} depends on the camera pose as well, it allows us to use the same loss to backpropagate and obtain gradients for pose refinement. 

As our goal is to refine the estimated noisy poses rather than perform SLAM, this training signal is sufficient for our needs. Moreover, we observe that poses tend to diverge with 3DGS due to the periodic opacity reset.
To combat this, we impose a soft constraint with an additional pose regularization, that encourages the matrices~$P^e_t$ to stay close to the identity matrix $I$:
\begin{equation}
    \mathcal{L}_{\text{pose}} = \| P^e_{t_{s_1}} - I \|_2 + \| P^e_{t_{s_2}} - I \|_2 + \| P^e_{t} - I \|_2
    \label{eq:L_pose} \mathrm{\,,}
\end{equation}
with all terms weighted equally.

\subsection{Optimization}
\label{ssec:Optimization} 

Eq.~\eqref{eq:L_recon_per_pixel} defines the reconstruction loss per pixel for a single event frame. However, naively averaging these per-pixel losses over whole images leads to problems. For small event windows, most pixels have no events, which are not very informative but will then make up the majority of the loss. 
To address this, we compute separate averages of the losses for pixels with events~$\mathcal{X}_\text{evs}$ and pixels without events~$\mathcal{X}_\text{noevs}$. 
These averages are then scaled by the hyperparameter~$\alpha=0.3$ to obtain the complete weighted reconstruction loss:
\begin{equation}
    \begin{split}
        \mathcal{L}_{\text{recon}}\left(t_s, t\right) = \,\,&
        \frac{\alpha}{|\mathcal{X}_{\text{noevs}}|} \cdot 
        \left(\sum_{\x\in \mathcal{X}_{\text{noevs}}} \mathcal{L}_{\x}\left(t_s, t\right)\right) + \\
        + \,\,& \,\, \frac{1 - \alpha}{|\mathcal{X}_{\text{evs}}|} \,\,\, \cdot 
        \left(\sum_{\x\in \mathcal{X}_{\text{evs}}} \mathcal{L}_{\x}\left(t_s, t\right)\right). 
    \end{split}
    \label{eq:L_recon}
\end{equation}
To obtain the final loss, we take a weighted sum of the reconstruction losses for the two event windows from Sec.~\ref{subsec:adaptive_window} along with the isotropic and pose regularization: 
\begin{equation}
    \begin{split}
        \mathcal{L} =\,\,\,\, & 
        \lambda_1 \mathcal{L}_{\text{recon}}\left(t_{s_1}, t\right) \,\,+  \,\,
        \lambda_2 \mathcal{L}_{\text{recon}}\left(t_{s_2}, t\right)  \\&
        +\,\, \lambda_\text{iso} \mathcal{L}_{\text{iso}} \,\,+ \,\,
        \lambda_\text{pose} \mathcal{L}_{\text{pose}}
    \end{split}
    \label{eq:loss} \mathrm{\,,}
\end{equation}
where $\lambda_1$, $\lambda_2$ and $\lambda_{\text{iso}}$ are hyper-parameters. In our experiments, we use  $\lambda_1=\lambda_2=0.65$, and $\lambda_{\text{iso}}$ is set to $10$ initially and reduced to $1$ after $\qty{d4}{}$ iterations.

\section{Experimental Evaluation}\label{sec:Experiments}

\subsection{Implementation details}

We provide the full implementation details in the supplemental material. Running our method on a scene takes one to two hours (depending on the scene size) with a single NVIDIA GeForce RTX 3090. 

\subsection{Datasets}
\label{sec:datasets}
We next describe the new event datasets we provide to analyze large-scale scenes, along with the existing datasets that we use in the experiments.  

\begin{figure}[t]
\centering
\begin{subfigure}{.5\linewidth}
\centering
\includegraphics[width=.95\linewidth]{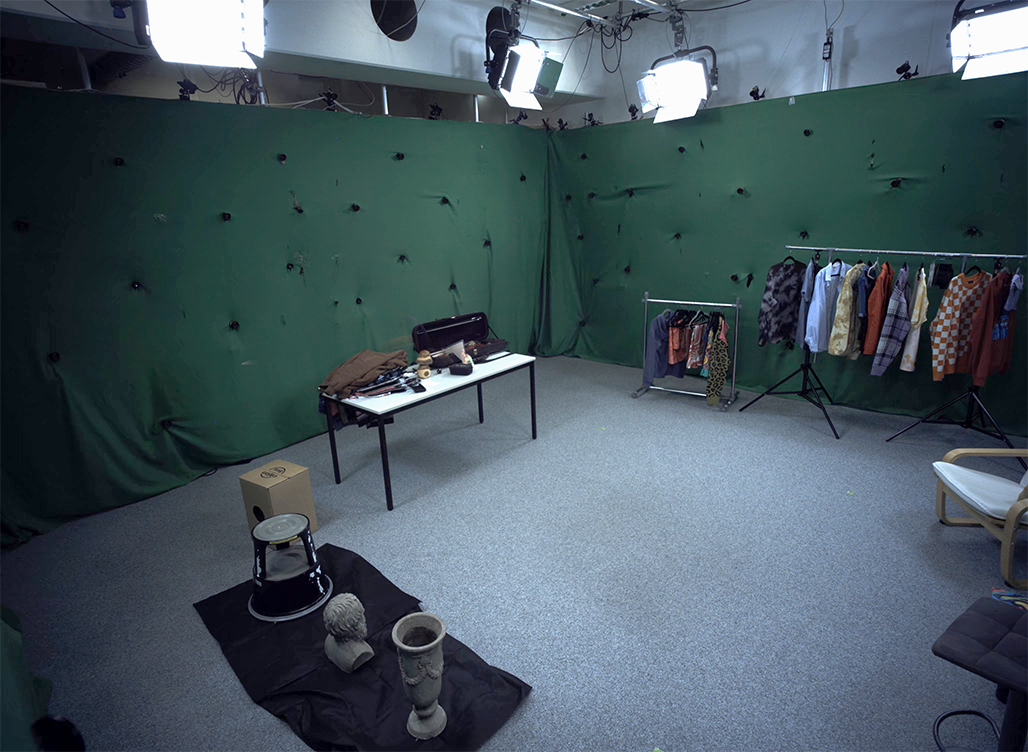}
\caption{} %
\label{fig:real_data_objects1}
\end{subfigure}%
\begin{subfigure}{.5\linewidth}
\centering
\includegraphics[width=.95\linewidth]{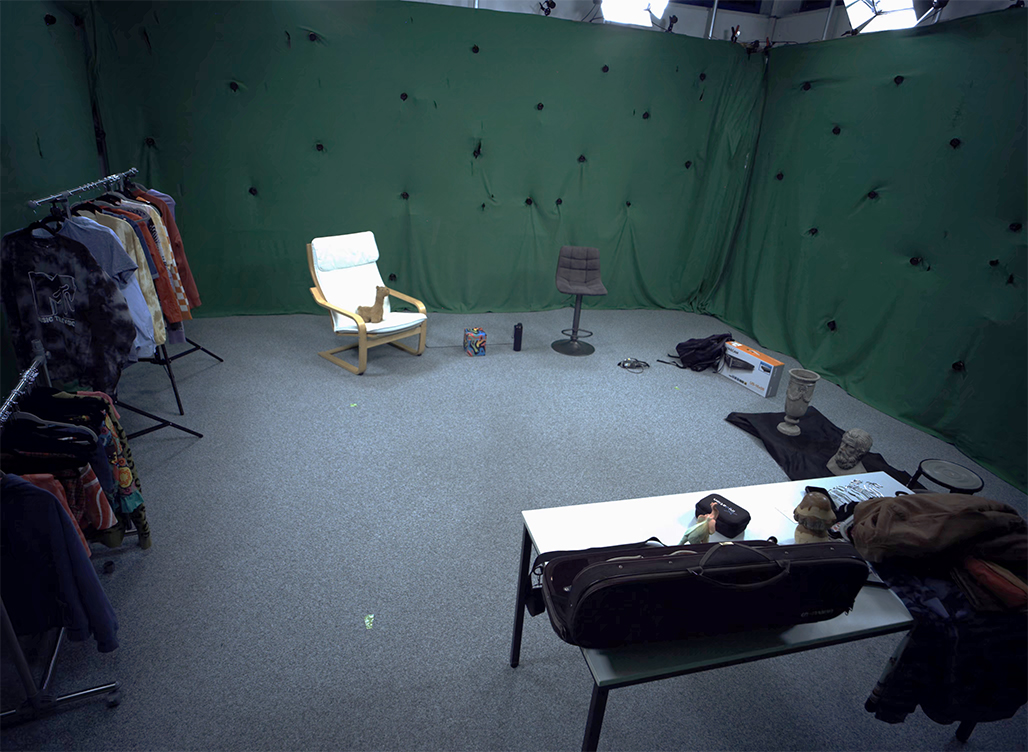}
\caption{} %
\label{fig:real_data_objects2}
\end{subfigure}
\caption{Two different views of the scene with inanimate objects assembled in the multi-view studio of MPI for Informatics. 
}
\label{fig:real_data_objects}
\end{figure}

\paragraph{E-3DGS-Real}
Our real dataset was captured within a studio environment. The scene consists of a diverse set of objects, as shown in Fig.~\ref{fig:real_data_objects}. We used a DAVIS346C color event camera to capture our scene with a resolution of $346 \times 260$. 
The contrast threshold settings were kept at their default values, which are symmetric. 
We capture multiple clips of the scene, each roughly $60{-}120\unit{\second}$ long with varying motion characteristics and levels of scene coverage. The captured data consists of the event stream and RGB images at 2.5 frames per second.
The studio is equipped with 115 traditional cameras distributed uniformly across the walls and capturing 4K footage at $\qty{50}{FPS}$.
Similar to the approach of Millerdurai et al.~and annotation of the EE3D-R (Real) dataset \cite{EventEgo3D}, we use these cameras to estimate and track the camera pose by detecting a checkerboard mounted to the event camera rig, providing tracking data at a frequency of up to $\qty{50}{Hz}$. Note that in some timestamps the checkerboard is not detected due to occlusions and thus the $\qty{50}{Hz}$ is only the best case. 
The data from the external cameras is relevant for camera pose estimation, but cannot be used as ground truth because of the significantly different perspectives from the training views.

\paragraph{E-3DGS-Synthetic}
For creating the synthetic dataset, we choose three scenes of UnrealEgo~\cite{unrealego}.
We rendered $60\unit{\second}$ clips of each scene at $\qty{1000}{FPS}$.
The scenes contain large-scale environments and exhibit various types of surfaces, including reflections.
We noticed that a few of the small highly reflective objects (e.g.,~metallic rods) cause unnatural aliasing in the renders, so we changed them to use diffuse materials. The event generation model from Sec.~\ref{eq:egm} was used to simulate event data from these high-fidelity frames. 
While we had access to pose data $\qty{1000}{Hz}$, we downsampled it to $\qty{50}{Hz}$ to simulate a real-world setting in which the poses are estimated from externally captured RGB images. 

\paragraph{E-3DGS-Synthetic-Hard}
This dataset is designed specifically to highlight and rigorously evaluate the key contributions of our method during the ablation study.
To assess the significance of our pose refinement module---which cannot be quantitatively evaluated on the E-3DGS-Real dataset---we introduce artificial noise into the E-3DGS-Synthetic dataset, which is carefully matched to the one observed in real data (see Supplement~\ref{sec:supp_dataset_perturb} for details). This allows us to assess the performance of our pose refinement module effectively. 
In addition to introducing noise, we also address the issue of camera speed variation. While the camera speed in the E-3DGS-Synthetic dataset generally stays within a narrow range, this does not fully test the capabilities of our adaptive event windows. To create a more challenging scenario, we varied the camera speed sinusoidally, with a ratio between its maximum and minimum speed of 100. This modification enables a more comprehensive evaluation of our adaptive event windows.

\paragraph{TUM-VIE}
This dataset consists of recordings from a Prophesee Gen4 sensor~\cite{klenk2021tum}. RGB views from an externally calibrated camera are also provided. The camera extrinsics are tracked at $\qty{120}{Hz}$. Two of the recordings have been used in Robust E-NeRF~\cite{robust_enerf}; we train our method on these recordings, namely \texttt{mocap-1d-trans} and \texttt{mocap-desk2} to compare with Robust E-NeRF. However, as also argued in Low and Lee \cite{robust_enerf}, these recordings are not well suited for novel view synthesis since the captures are predominantly front-facing, with some small displacements either in circles or from side to side. 

\paragraph{EventNeRF Datasets}
EventNeRF \cite{eventnerf} provides $\ang{360}$ object-centric event data, which we use to show that our method also outperforms previous methods on object-centric data. To be consistent with the original work, we evaluate our method on poses that are a part of the training trajectory instead of novel views, for our evaluation metrics to be comparable to theirs. We train our method on the synthetic sequences to perform the quantitative comparison. In these experiments, the background color is set to $159/255$, following the original paper \cite{eventnerf}.

\begin{figure*}[h!]
\centering
\vspace{-10pt}
\includegraphics[width=1.0\linewidth]{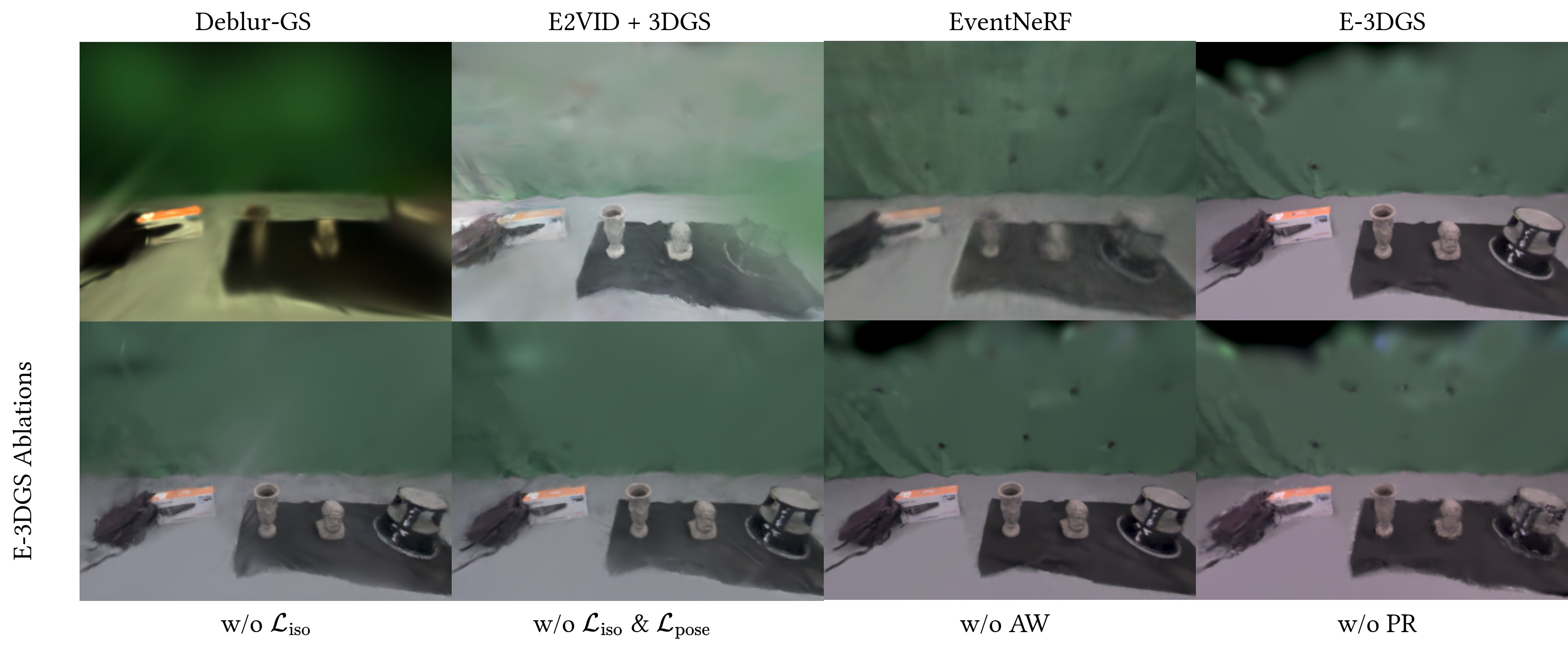}
\caption{Comparison of E-3DGS against the baselines and ablation study on the E-3DGS-Real dataset. Deblur-GS, E2VID + 3DGS and EventNeRF suffer from various issues including blurring, floaters, and noise. In contrast, our method delivers clear details, such as the intricate structure of the sculpture's face. } 
\vspace{-5pt}
\label{fig:real_data_ablation}
\end{figure*}

\setlength{\tabcolsep}{6.9pt}

\begin{table*}[t]
  \centering
  \sisetup{detect-family=true, text-series-to-math = true, propagate-math-font = true}
  \resizebox{\linewidth}{!}{
  \begin{tabularx}{1.3\linewidth}{p{0.205\hsize} |S S S| S S S| S S S || S S S}
    \toprule
    \multirow{2}{*}{Method} & \multicolumn{3}{c|}{Company}  & \multicolumn{3}{c|}{ScienceLab} & \multicolumn{3}{c||}{Subway} & \multicolumn{3}{c}{Average}\\
    \cline{2-13}
    & $\uparrow$ PSNR & $\downarrow$ LPIPS & $\uparrow$ SSIM & 
    $\uparrow$ PSNR & $\downarrow$ LPIPS & $\uparrow$ SSIM & 
    $\uparrow$ PSNR & $\downarrow$ LPIPS & $\uparrow$ SSIM & 
    $\uparrow$ PSNR & $\downarrow$ LPIPS & $\uparrow$ SSIM \\
    \hline
    
    EventNeRF \cite{eventnerf} &    
    \2 19.59	& \3 0.41	& \2 0.65	& 
    \2 17.22	& \3 0.46	& \2 0.60	& 
    \2 18.71	& \3 0.34	& \2 0.67  & 
    \2 16.80	& \3 0.50	& \2 0.61  \\
    
    E2VID \cite{e2vid} + 3DGS \cite{3dgs} &    
    \3 9.79  & \2 0.37 & \3 0.48	  &    
    \3 11.86 & \2 0.38 & \3 0.54   &     
    \3 9.79  & \2 0.40 & \3 0.43    &     
    \3 10.48 & \2 0.38 & \3 0.49 \\
    
    E-3DGS (ours) &    
    \1 20.78   &   \1 0.29   &   \1 0.72   &
    \1 18.41   &   \1 0.28   &   \1 0.73   &
    \1 19.92   &   \1 0.20   &   \1 0.74   &
    \1 19.70   &   \1 0.26   &   \1 0.73
    \\
  \bottomrule
  \end{tabularx}
  }
  \vspace{-5pt}
 \caption{
      Comparison of several methods on the E-3DGS-Synthetic dataset: We outperform the baselines by a large margin in all cases. Furthermore, E2VID + 3DGS shows lower PSNR but achieves better LPIPS than EventNeRF due to E2VID's frame reconstruction, which has poor color consistency but an adequate level of edge details (see Fig.~\ref{fig:synthetic_e3dgs}).
      Green and yellow are the best and the second-best, respectively. 
  }
    \label{tab:comparison_synthetic}
\end{table*}

\setlength{\tabcolsep}{\orgtabcolsep}

\setlength{\tabcolsep}{4.1pt}

\begin{table}[t]
  \centering
  \sisetup{detect-family=true, text-series-to-math = true, propagate-math-font = true}
  \resizebox{1.0\columnwidth}{!}{
  \begin{tabularx}{1.2\columnwidth}{p{0.2\columnwidth} | S S S| S S S}
    \toprule
    \multirow{2}{*}{Scene} & \multicolumn{3}{c|}{EventNeRF \cite{eventnerf}} & \multicolumn{3}{c}{E-3DGS (ours)}\\
    \cmidrule{2-7}
    &$\uparrow$ PSNR & $\downarrow$ LPIPS & $\uparrow$ SSIM
    &$\uparrow$ PSNR & $\downarrow$ LPIPS & $\uparrow$ SSIM\\
    \midrule 
    Chair & 
    \1 30.62  &  \2 0.05  &  \2 0.94 & 
    \2 30.42  &  \1 0.03  &  \1 0.95\\
    Drums & 
    \2 27.43  &  \2 0.07  &  \2 0.91 & 
    \1 31.07  &  \1 0.03  &  \1 0.95\\
    Ficus & 
    \2 31.94  &  \2 0.05  &  \2 0.94 & 
    \1 34.08  &  \1 0.02  &  \1 0.96\\
    Hotdog & 
    \2 30.26  &  \2 0.04  &  \2 0.94 & 
    \1 30.79  &  \1 0.03  &  \1 0.96\\
    Lego & 
    \2 25.84  &  \2 0.13  &  \2 0.89 & 
    \1 30.74  &  \1 0.04  &  \1 0.94\\
    Materials & 
    \2 24.10  &  \2 0.07  &  \2 0.94 & 
    \1 33.73  &  \1 0.02  &  \1 0.97\\
    Mic & 
    \2 31.78  &  \2 0.03  &  \2 0.96 & 
    \1 35.87  &  \1 0.02  &  \1 0.98\\
    \midrule
    Average & 
    \2 28.85  &  \2 0.06  &  \2 0.93 & 
    \1 32.39  &  \1 0.03  &  \1 0.96\\
  \bottomrule
  \end{tabularx}
  }
 \caption{
    Comparisons on the synthetic EventNeRF dataset. Our method demonstrates significant improvements over EventNeRF across all evaluation metrics.
  }
  \label{tab:eventnerf_synthetic}
\end{table}

\setlength{\tabcolsep}{\orgtabcolsep}

\subsection{Evaluation Metrics} 
For E-3DGS-Real dataset, the RGB frames are of too low quality to be used for evaluation purposes, and, therefore, we only perform qualitative comparisons. 
With TUM-VIE, as suggested in Robust E-NeRF~\cite{robust_enerf}, it is not trivial to do the tone mapping correctly. 
Therefore, we do quantitative evaluation only with the synthetic datasets. 
For the evaluation on synthetic data, keeping in line with the previous literature, we adopt the following evaluation metrics: 
\begin{itemize} 
    \item Peak Signal-to-Noise Ratio (PSNR); 
    \item Learned Perceptual Image Patch Similarity (LPIPS) \cite{LPIPS}; 
    \item Structural Similarity Index Measure (SSIM). 
\end{itemize}

\setlength{\tabcolsep}{8.5pt}

\begin{table*}[t]
  \centering
  \resizebox{\linewidth}{!}{
  \begin{tabularx}{1.47\linewidth}{c c c c| c c c| c c c| c c c|| c c c}
    \toprule
    \multicolumn{4}{c|}{Components} & \multicolumn{3}{c|}{Company}  &  \multicolumn{3}{c|}{ScienceLab} & \multicolumn{3}{c||}{Subway} & \multicolumn{3}{c}{Average} \\
    \hline
    $\mathcal{L}_\text{iso}$ & $\mathcal{L}_\text{pose}$ & PR & AW &  
    $\uparrow$ PSNR & $\downarrow$ LPIPS & $\uparrow$ SSIM & 
    $\uparrow$ PSNR & $\downarrow$ LPIPS & $\uparrow$ SSIM & 
    $\uparrow$ PSNR & $\downarrow$ LPIPS & $\uparrow$ SSIM & 
    $\uparrow$ PSNR & $\downarrow$ LPIPS & $\uparrow$ SSIM \\
    \hline
 \cxmark & \cxmark & \cxmark & \cxmark     &     
 \2 20.742  &  \1 0.404  &  \1 0.661     &      
 \1 18.823  &  \1 0.414  &  \1 0.677     &      
 \3 18.923  &  \1 0.436  &  \1 0.619     &       
 \1 19.496  &  \1 0.418  &  \1 0.652 \\
\cdashline{1-4}

  & \cxmark & \cxmark & \cxmark     &     
   20.519  &  \3 0.434  &  \3 0.631      &      
  \3 18.099  &  \3 0.454  &  \3 0.631     &      
  \2 19.401  &   0.475  &  \3 0.601     &       
  \3 19.340  &  \3 0.454  &  \3 0.621 \\
 \cdashline{1-4}

 \cxmark & \cxmark & & \cxmark     &     
 20.229 &  0.539  &  0.606      &      
 17.646  &  0.587  &  0.601     &      
 18.746  &  0.620  &  0.569     &       
 18.874  &   0.582  &   0.592 \\
 \cdashline{1-4}

 \cxmark & \cxmark & \cxmark &      &     
 \3 20.667  &  \2 0.427  &  \2 0.642       &      
 \2 18.354  &  \2 0.440  &  \2 0.657      &      
 18.742  &  \2 0.440  &  \2 0.606      &       
 19.254  &  \2 0.436  &  \2 0.635 \\
 \cdashline{1-4}
 
 & \cxmark & \cxmark &      &     
 \1 20.845 &  0.441  &  0.623      &      
 17.792  &  0.472  &  0.616     &     
 \1 19.475  & \3 0.469  &  0.600     &      
 \2 19.371  &   0.460  &   0.613 \\
 \cdashline{1-4}

 & & \cxmark & \cxmark      &  
 19.834 &  0.537  &  0.583       &    
 17.317  &  0.577  &  0.571      &    
 18.111  &  0.605  &  0.532      &  
 18.421  &   0.573  &   0.562 \\
  \bottomrule
  \end{tabularx}
  }
  \caption{
    Ablation study on the E-3DGS-Synthetic-Hard dataset. The overall tendency is that the performance declines when one of the components is removed, confirming their contribution to the overall performance. Notably, E-3DGS without AW consistently ranks second, while omitting $L_{\text{iso}}$ often results in third place or close.
    (PR: \textbf{P}ose \textbf{R}efinement, AW: \textbf{A}daptive Event \textbf{W}indow). Green, yellow, and orange indicate the best, second-best, and third-best results, respectively. 
  }
  \label{tab:ablation_study}
\end{table*}

\setlength{\tabcolsep}{\orgtabcolsep}

\subsubsection{Color Correction} 
As our method only learns logarithmic differences rather than absolute color intensities, there is an ambiguity in the reconstructed color balance and illumination of the scene. Hence, color needs to be adjusted, as otherwise, the evaluation metrics will be less meaningful. We correct predicted images using the following equation: 
\begin{equation}
    L'_c = L' + (\mathbb{E}[L] - \mathbb{E}[L']) \mathrm{\,,}
    \label{eq:color_correction}
\end{equation}
where $L'_c$ is the color corrected logarithmic image and ``$\mathbb{E}[\cdot]$'' is the expectation operator. 
Eq.~\eqref{eq:color_correction} is applied separately to each color channel, which effectively aligns the per-channel logarithmic means of the predicted images with the ground-truth ones. 
Since in the synthetic setting, we already know the exact contrast threshold, there is no need for correcting the scale of the image as done in some previous works~\cite{eventnerf,robust_enerf}. 
Since we lack reference images for the real dataset, neither evaluation nor color correction is applicable to it. However, some minor color and contrast adjustments are manually made for better visualization. 

\subsection{Comparisons to Related Methods}
\label{subsec:comparisons}

\paragraph{RGB-Based Methods} 
We train Deblur-GS~\cite{deblurgs} on blurry RGB images from our E-3DGS-Real dataset to establish a reference using RGB inputs. 
We also convert the event stream to images using E2VID~\cite{e2vid} and apply 3DGS (referred to as ``E2VID + 3DGS''). 
This method is evaluated on all E-3DGS datasets. 
To train both methods, we interpolate the camera poses at discrete time steps provided by the external tracking system, which is necessary because the pose timestamps do not align with the frame timestamps. 
We use Spherical Linear Interpolation (SLERP) for the rotations and Linear Interpolation (LERP) for the translations to obtain the camera poses for the images. 

\paragraph{Event-Based Methods}
For comparison with event-based methods, we train EventNeRF~\cite{eventnerf} on all E-3DGS datasets. To adapt it for our datasets, we normalize the camera poses within a unit sphere and following NeRF++~\cite{nerf++} added a background network to model areas outside the sphere, as the scene extent is unknown. Furthermore, the maximum event window length is increased by the factor of $10$ to aid convergence (up to one second). 
We do not train our method on the synthetic dataset provided by Robust E-NeRF~\cite{robust_enerf}, as it is designed for extremely long refractory periods that are not observed in other datasets. However, we compare their method to ours on two sequences from TUM-VIE in Fig.~\ref{fig:real_data_ablation}, namely \texttt{mocap-1d-trans} and \texttt{mocap-desk2}.

\subsubsection{Observations}

\begin{figure}[ht]
\centering
\includegraphics[width=1.0\columnwidth]{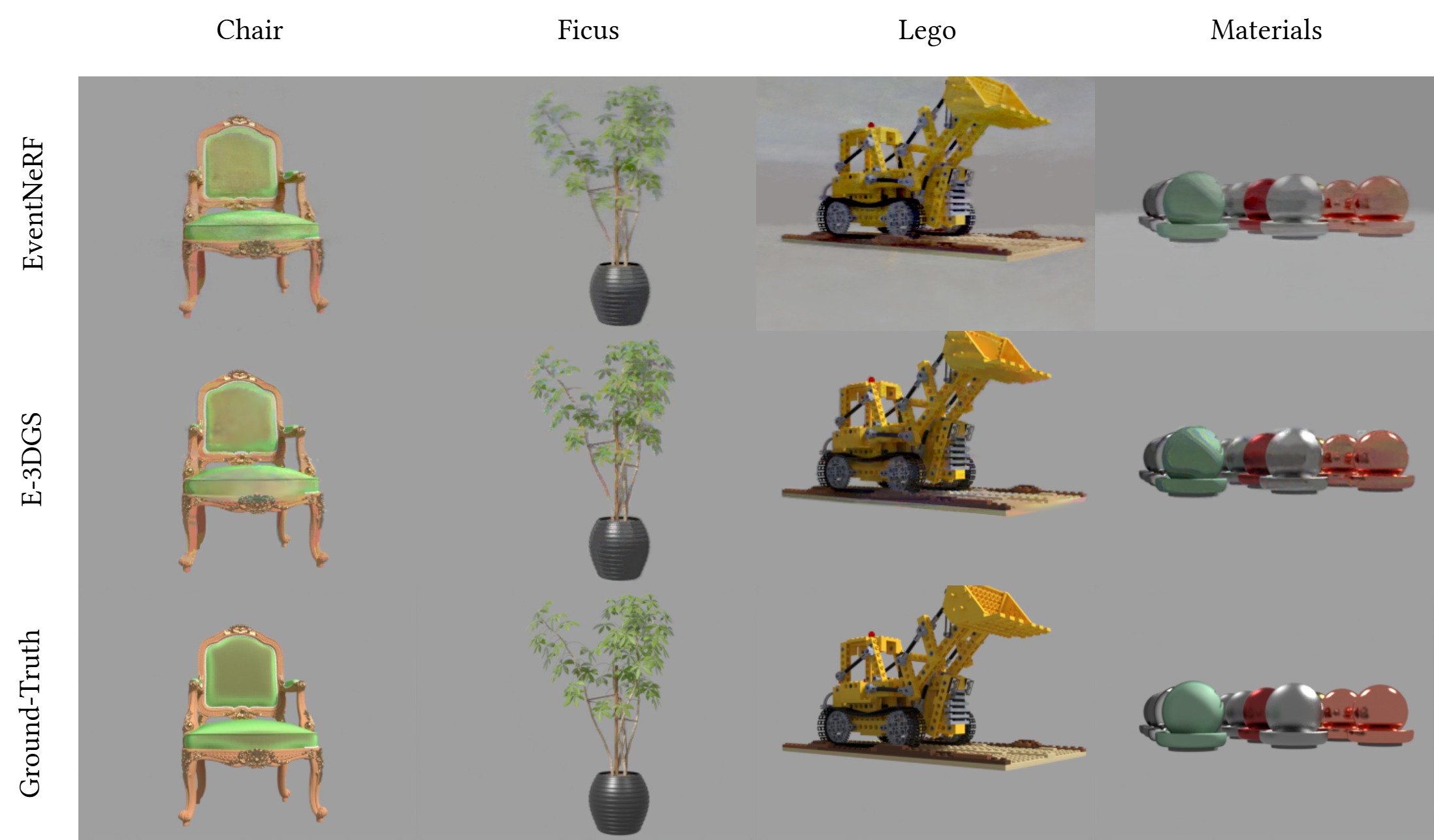}
\caption{
Comparison of E-3DGS vs. EventNeRF on the synthetic EventNeRF dataset. EventNeRF struggles with noise in the Drums sequence, blurriness in Ficus, and background artifacts in Lego and Materials sequences, while E-3DGS handles these issues well.}
\label{fig:synthetic_eventnerf}
\end{figure}

\begin{figure}[t]
\centering
\includegraphics[width=\columnwidth]{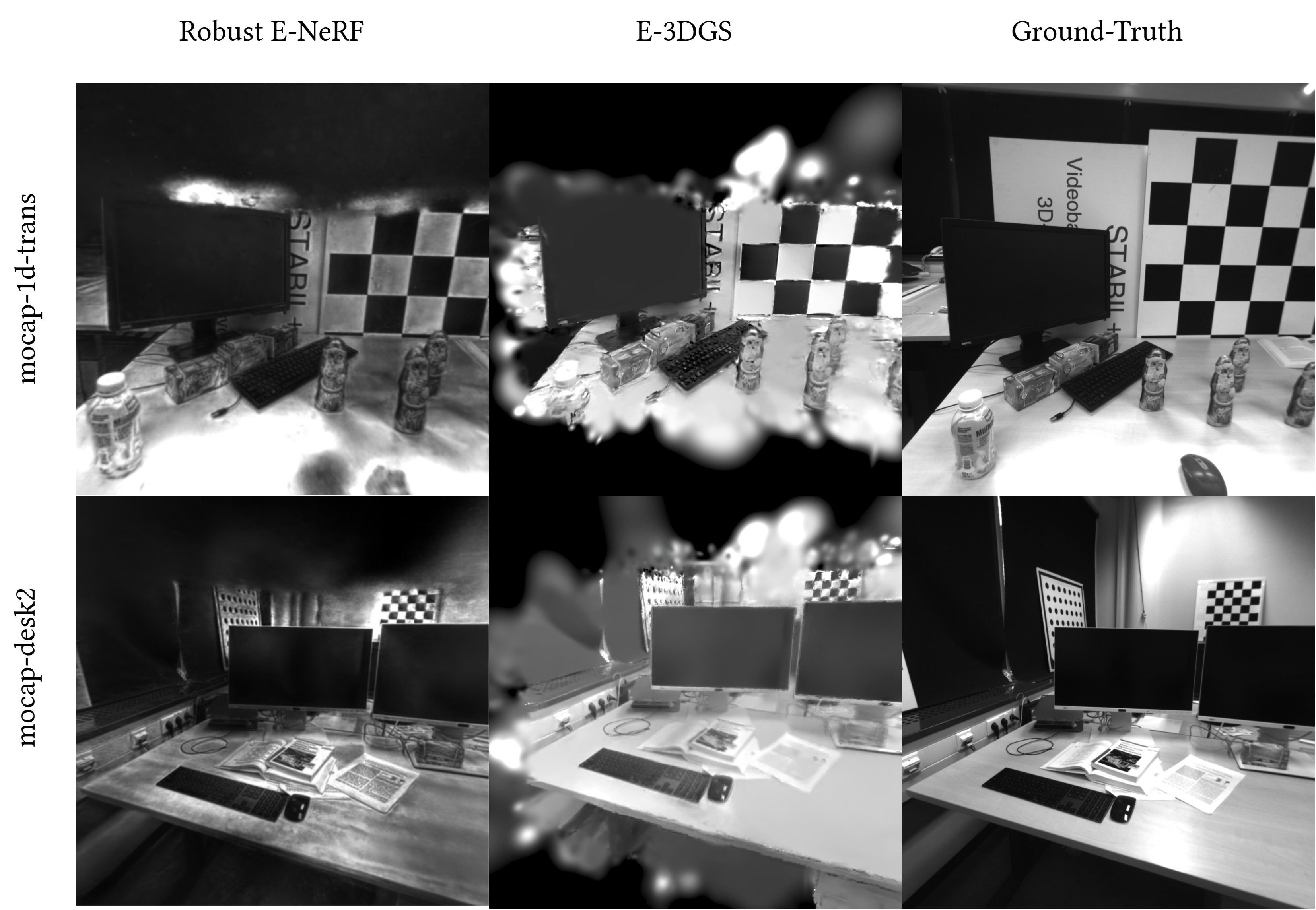}
\caption{Comparison of E-3DGS vs.~Robust E-NeRF on the TUM-VIE dataset.
While Robust E-NeRF achieves higher local contrast, it suffers from globally inconsistent brightness. E-3DGS produces consistent brightness across the scene, albeit with some detail loss (e.g.,~in the table texture of the mocap-desk2 sequence).
} 
\label{fig:tumvie_comp}
\end{figure}

\begin{figure}[t]
\centering
\includegraphics[width=\columnwidth]{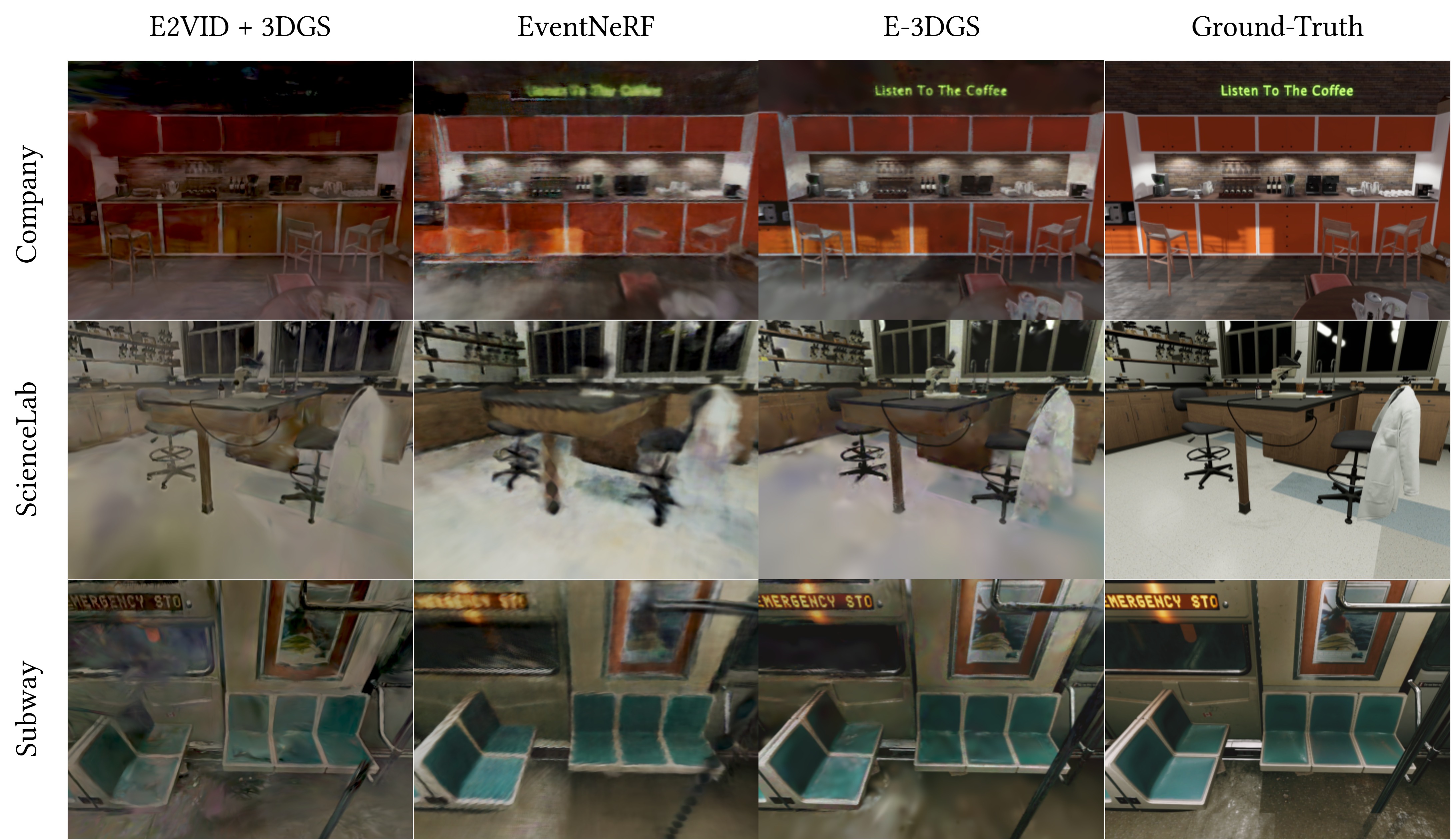}
\caption{Comparison of E-3DGS vs.~baselines on the E-3DGS-Synthetic dataset. E2VID + 3DGS struggles with poor color reconstruction but captures edges and structure reasonably well. EventNeRF suffers from noise and a lack of sharpness. In contrast, our method delivers clear details and accurate colors, with only minor issues in certain areas (such as the coat on a chair in the ScienceLab sequence. Best viewed with zoom.}
\label{fig:synthetic_e3dgs}
\end{figure}

The results of all evaluations are reported in Tables \ref{tab:comparison_synthetic}--\ref{tab:eventnerf_synthetic} and Figs.~\ref{fig:real_data_ablation}--\ref{fig:synthetic_e3dgs}. 
As visible, our method consistently outperforms the baselines both on synthetic and real data. 
In the EventNeRF object-centric datasets, our method shows clear superiority across almost all evaluation metrics. The only exception is a marginally lower PSNR score on the ``Chair" scene, as detailed in Table~\ref{tab:eventnerf_synthetic}. The general performance advantage is further backed by the qualitative results in Fig.~\ref{fig:synthetic_eventnerf}, where our method produces more accurate reconstructions.

Similarly, on the E-3DGS-Synthetic dataset, E-3DGS significantly surpasses both EventNeRF and E2VID+3DGS by a wide margin; see Table~\ref{tab:comparison_synthetic}. The qualitative results on the E-3DGS-Real dataset, highlighted in Fig.~\ref{fig:real_data_ablation}, further demonstrate our method's superior performance: Deblur-GS struggles with excessive blur; EventNeRF suffers from noise due to ray sampling and memory constraints, and E2VID+3DGS exhibits noisy Gaussians and floaters. 

While Robust E-NeRF achieves higher local contrast, it struggles with global brightness consistency due to single-event training; see Fig.~\ref{fig:tumvie_comp}. Our E-3DGS maintains consistent brightness across the scene, with only a slight reduction in local contrast. 
Note that we can observe some holes and floaters near the outer peripheries in Figs.~\ref{fig:real_data_ablation} and~\ref{fig:tumvie_comp}. These effects are due to out-of-bound areas at the edges of the observations that occur as a result of the undistortion of the event stream.

\subsection{Ablation Studies}

To evaluate the effects of individual contributions, we do extensive qualitative and quantitative ablation studies. 
We primarily train different variants of our method on the E-3DGS-Real and E-3DGS-Synthetic-Hard datasets, focusing on the effects of four key components: $\mathcal{L}_\text{iso}$, $\mathcal{L}_\text{pose}$, \textbf{P}ose \textbf{R}efinement (PR), and the \textbf{A}daptive Event \textbf{W}indow (AW).

For the ablation experiments without adaptive window, we use a maximum time interval $T_\text{max}$ instead of maximum events $N_\text{max}$ to sample the event windows. The value of $T_\text{max}$ is computed from $N_\text{max}$, such that the average event window size remains approximately similar.

The results are reported in Table~\ref{tab:ablation_study} and Figs.~\ref{fig:real_data_ablation} and~\ref{fig:synthetic_e3dgs}.
Removing $L_{iso}$ results in a noticeable performance drop, but removing $L_{iso}$ and $L_{pose}$ jointly leads to a much more significant decline. 
This is likely because $L_{pose}$ prevents pose divergence in unstable conditions, while removal of $L_{iso}$ causes instability due to overfitting. 
Similar effects could occur when combining $L_{iso}$ with pose refinement.

\section{Conclusion}\label{sec:Conclusion} 
\vbox{
We show that E-3DGS effectively combines the strengths of 3D Gaussian splatting and event-based supervision for 3D reconstruction and novel view synthesis of large-scale scenes. 
It significantly outperforms the baselines quantitatively and qualitatively, while being orders of magnitude faster. 
One aspect beyond the scope of this paper 
is lifting the requirement for camera pose initialization through an external process. 
We believe this work paves the way for robust and scalable large-scale scene reconstruction utilizing the advantages of event cameras to capture details in challenging conditions, such as low light and fast motion. 
}

{
    \small
    \bibliographystyle{ieeenat_fullname}
    \bibliography{main}
}
\clearpage
\maketitlesupplementary
\setcounter{figure}{0}
\renewcommand{\thefigure}{\Roman{figure}}
\setcounter{section}{0}
\renewcommand{\thesection}{\Roman{section}}
\vspace{15pt}

This supplement provides additional details and insights into the methods and experiments discussed in the main paper.
In Sec.~\ref{sec:supp_frustum}, we elaborate on our frustum-based initialization, explaining the sampling strategy and how it ensures effective Gaussian placement in the scene.
Sec.~\ref{sec:supp_pose_refinement} provides further details on our pose refinement, specifically the use of Gram-Schmidt orthogonalization to maintain valid transformations during optimization. 
In Sec.~\ref{sec:supp_dataset_perturb}, we analyze the camera pose noise in the E-3DGS-Real dataset and describe the process we use to simulate realistic pose perturbations for the E-3DGS-Synthetic-Hard dataset.
Sec.~\ref{sec:supp_implementation_details} outlines the implementation details, including adjustments to the original 3DGS training schedule to improve convergence.
Sec.~\ref{sec:supp_evaluation} covers our evaluation, highlighting the measures we take to ensure reliable results, particularly for the ablation studies. 
Finally, we present a comprehensive comparison in Sec.~\ref{sec:additional_comparisons}  showcasing additional visual results and ablation studies on the E-3DGS-Real, E-3DGS-Synthetic, and E-3DGS-Synthetic-Hard datasets. 
These experiments expand on the results from the main paper and further demonstrate the effectiveness of our method across different scenarios. 

\section{Frustum-Based Initialization}
\label{sec:supp_frustum}

As described in Sec.~\ref{sec:frustum_init} of the main paper, our approach involves initializing a fixed number of Gaussians, denoted as $N_g$. If we have $N_t$ camera poses, we distribute the Gaussians across these poses, resulting in $N_g / N_t$ Gaussians being initialized for each pose.
The initialization process begins by sampling points within the camera frustum in normalized device coordinates (NDC). However, instead of uniformly sampling all three coordinates $(x, y, z)$ in NDC, we adopt a different strategy for depth (z-axis).

We observe that when depth was sampled directly in NDC, most Gaussians would cluster very close to the near plane ($z_\text{near}$), leading to poor scene coverage. To address this, we sample the depth uniformly in camera coordinates between $z_\mathrm{near}$ and $z_\mathrm{far}$. This ensures a more even distribution of Gaussians across the entire depth range.

Once the depth is sampled in camera coordinates, it is converted into NDC. Next, the $x$ and $y$ coordinates are sampled uniformly in NDC. With $x$, $y$, and $z$ values now in NDC, we un-project them back into the world coordinates. This conversion gives us the final positions for the Gaussians in the 3D scene.
Next, the entire process is repeated for each camera frustum associated with the given poses $P_t$, ensuring a comprehensive initialization across all views. Therefore, the distribution of Gaussians is effectively tied to the observable scene regions. 

\section{Pose Refinement and Gram-Schmidt Orthogonalization}
\label{sec:supp_pose_refinement}

In Sec.~\ref{sec:pose_refinement} of the main paper, we introduce our approach to pose refinement, where the refined pose $P'_t$ is modeled as $P'_t = P^e_t P_t$, with $P^e_t$ being an error correction transform. Rather than directly optimizing $P^e_t$ as a $3{\times}3$ matrix, we represent it using two rotation vectors $r_1$ and $r_2$ and a translation vector $T$, following the method of Hempel et al.~\cite{6d_rotation}. This representation allows us to ensure that $P^e_t$ remains a valid transformation matrix during optimization. 

To maintain the orthogonality of the rotation matrix, we apply Gram-Schmidt orthogonalization to $r_1$ and $r_2$ to compute the final rotation matrix $R = [r'_1, r'_2, r'_3]$. The process is as follows:

\allowdisplaybreaks
\begin{equation}
\begin{aligned}
    r'_1 &= \frac{r_1}{\|r_1\|}, \\
    r'_2 &= \frac{r_2 - (r'_1 \cdot r_2)r'_1}{\|r_2 - (r'_1 \cdot r_2)r'_1\|},\\
    r'_3 &= r'_1 \times r'_2, \,\text{and}\\
    P^e_t &= 
    \begin{bmatrix}
    | & | & | & |\\
    r'_1 & r'_2 & r'_3 & T\\
    | & | & | & |\\
    0 & 0 & 0 & 1\\
    \end{bmatrix}.    
\end{aligned}
\end{equation}

Here, $r'_1$ is the normalized version of $r_1$, and $r'_2$ is obtained by subtracting the projection of $r_2$ onto $r'_1$ and normalizing the result. The third vector $r'_3$ is calculated as the cross product of $r'_1$ and $r'_2$, ensuring that the resulting rotation matrix is orthogonal. The final error correction matrix $P^e_t$ is then constructed using these orthogonal vectors and the translation vector $T$.

This approach guarantees that the pose refinement remains valid throughout the optimization process, contributing to the stability and accuracy of our method.

\begin{figure*}[!ht]
\centering
\begin{subfigure}[b]{0.8\linewidth}
    \centering
    \includegraphics[width=\linewidth]{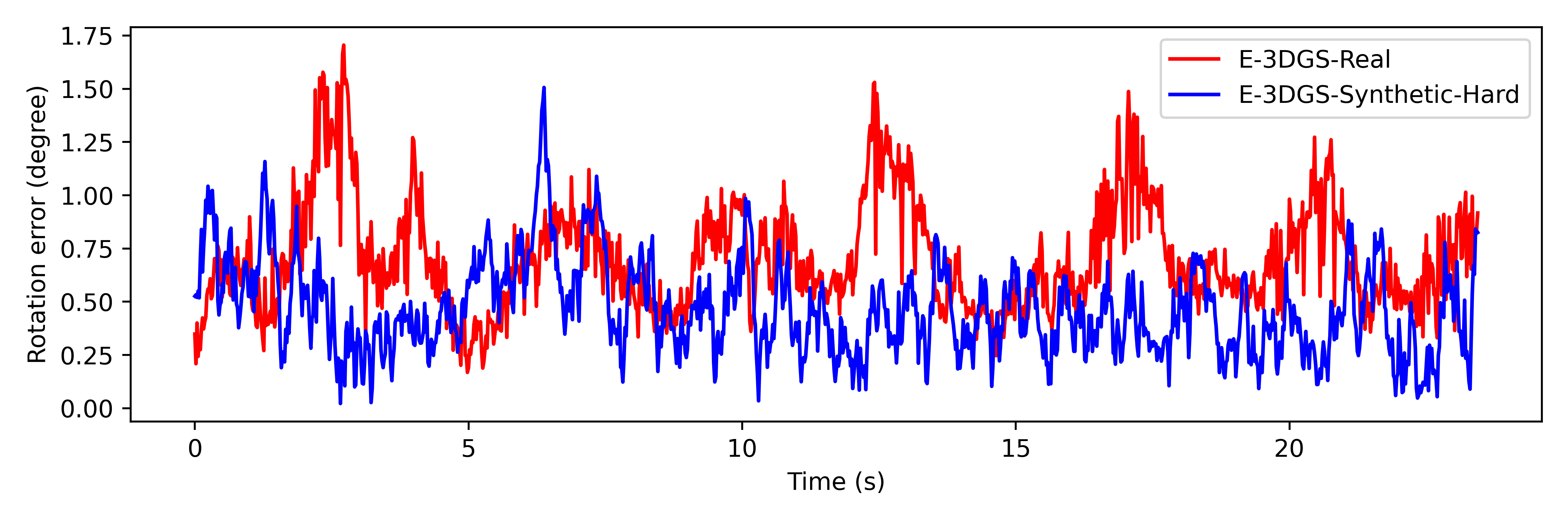}
    \caption{Rotation errors for both E-3DGS-Real and E-3DGS-Synthetic-Hard show a similar error distribution.}
\end{subfigure}
\hfill
\begin{subfigure}[b]{0.8\linewidth}
    \centering
    \includegraphics[width=\linewidth]{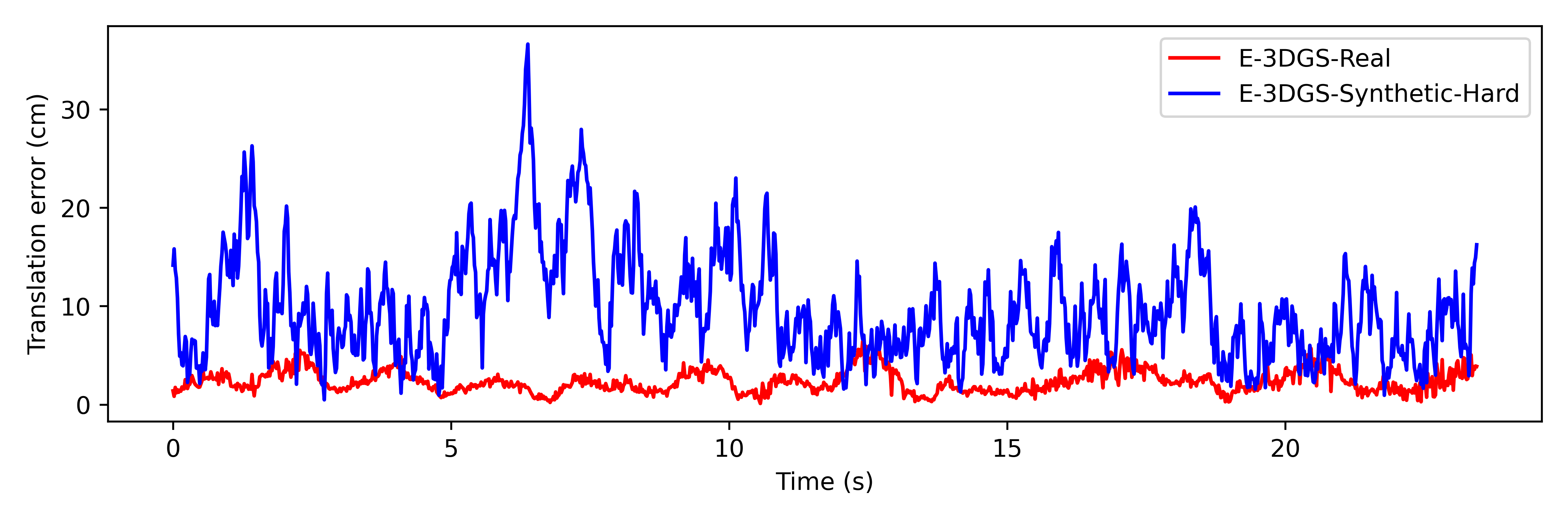}
    \caption{Larger translation errors are applied to E-3DGS-Synthetic-Hard, compared to those in E-3DGS-Real, to account for the larger scene size and ensure a sufficiently challenging difficulty level for meaningful ablation studies.}
\end{subfigure}
\caption{Comparison of estimated pose errors in the E-3DGS-Real dataset versus the synthetically introduced errors in the E-3DGS-Synthetic-Hard dataset. The synthetic perturbations are generated using an Ornstein–Uhlenbeck process to match the time-correlated nature and variance of the real data.}
\label{fig:noise}
\end{figure*}

\section{Pose Perturbation in E-3DGS-Synthetic-Hard}
\label{sec:supp_dataset_perturb}

As described in Sec.~\ref{sec:datasets} of the main paper, we provide the E-3DGS-Synthetic-Hard dataset that differs from E-3DGS-Synthetic in two aspects: 1) The camera speed is highly varied and 2) the camera extrinsics exhibit noise similar in characteristics to the noise observed in the real data. 
To quantify the camera pose noise in the E-3DGS-Real dataset, we compare the refined training camera trajectories with the initial trajectories. Our analysis reveals that these errors are time-correlated. Based on this observation and by examining the scale of these errors, we introduce synthetic perturbations in the E-3DGS-Synthetic dataset using a random walk with decay, specifically the Ornstein–Uhlenbeck process~\cite{pavliotis2014stochastic}, which ensures the perturbations have zero mean while remaining time-correlated. 

We calibrate the variance of the synthetic perturbations to match the rotation errors observed in the real data. For translation, we apply a higher level of perturbation, given that the synthetic scenes are significantly larger in scale than the real data. This adjustment ensures that translation errors are proportionally scaled, creating a comparable difficulty level for the ablation studies. The noise patterns are illustrated in Fig.~\ref{fig:noise}.

\section{Implementation Details}
\label{sec:supp_implementation_details}

Our codebase is based on 3DGS~\cite{3dgs}. We train the method for $\qty{6d4}{}$ instead of $\qty{3d4}{iterations}$, allowing the pose refinement to converge. 
The original paper performs both, densification and opacity resets of the Gaussians until $\qty{1.5d4}{iterations}$.
In our case, we perform opacity resets until $\num{3d4}$ and densification until $\qty{5d4}{iterations}$. 
From our analysis---while opacity resets are important to remove floaters---they also hamper the reconstruction quality. 
Therefore, once the scene is reasonably converged, we stop resetting opacity and only densify the scene to get better reconstruction.

Furthermore, 3DGS uses the fixed threshold value $\num{2d-4}$ to decide whether a Gaussian should be split up during the densification.
We start the optimization with the same value,  however, we linearly decrease it to $\num{4d-5}$ over ${\qty{4d4}{iterations}}$.
First, this allows our method to refine the poses with larger Gaussians, providing more support, and second, reduce the threshold in later stages to obtain a more detailed reconstruction.
We initialize $N_g={\qty{5d4}{}}$ Gaussians in all our trainings.

In the experiments with pose refinement, we restrict the number of spherical harmonics to one, as it allows for better pose refinement \cite{3dgsslam, splatam}. 
For the experiments with perfect poses, we follow the original 3DGS approach and use three spherical harmonics. 
In all experiments, except those conducted with the EventNeRF dataset \cite{eventnerf}, we consistently use $N_\text{max}{=}\num{e6}$ events for the window size. 
As sequences of the latter are very short and do not contain enough events for such large windows, we use $N_\text{max}{=}\num{e5}$ for them. 
Training the full method takes one to two hours with a single NVIDIA GeForce RTX 3090, depending on the scene size. 

\begin{figure*}[h]
\centering
\includegraphics[height=0.9\textheight]{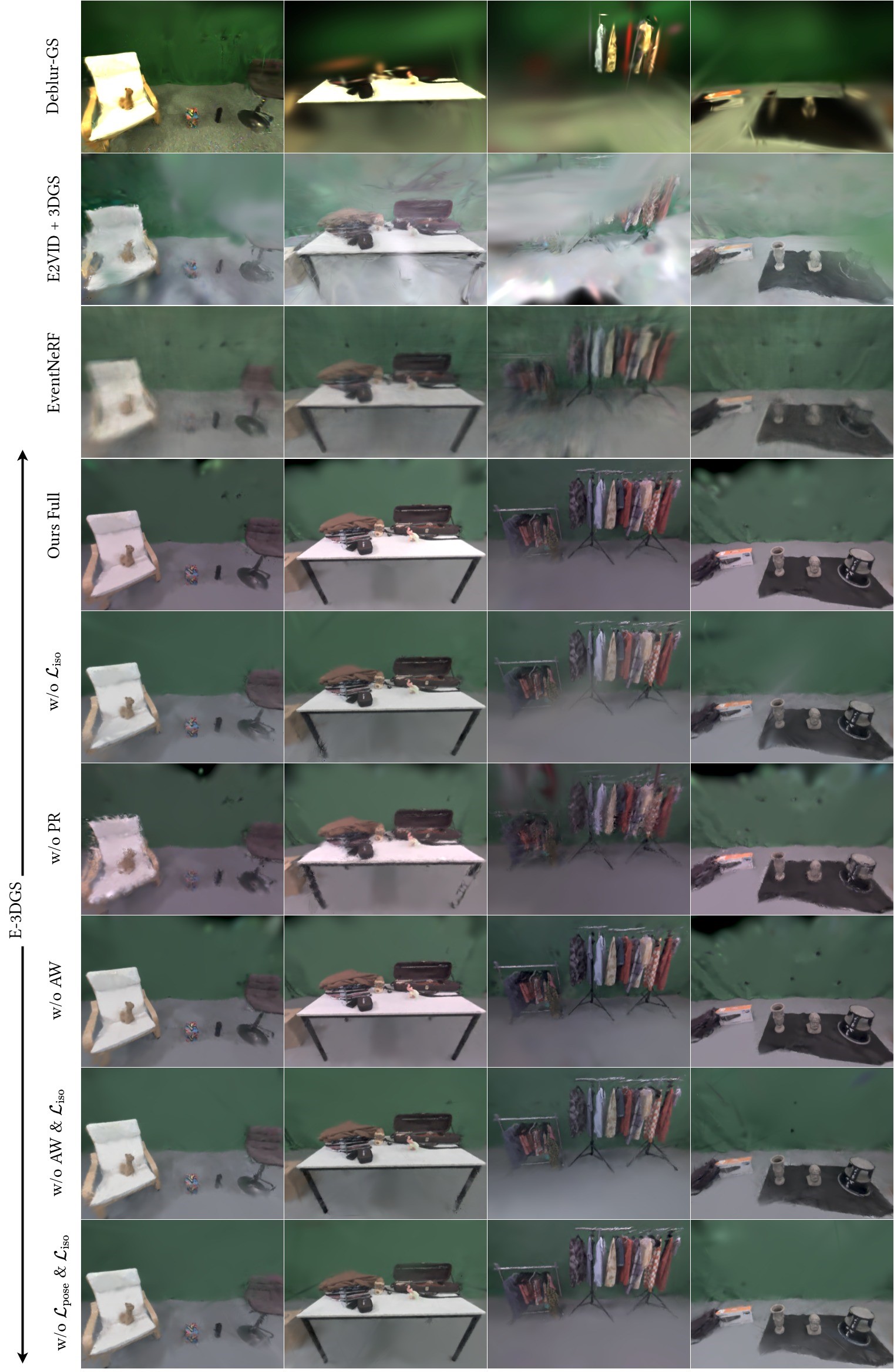}
\caption{Comparison of E-3DGS against the baselines and ablation study on E-3DGS-Real. As observed in the main paper, Deblur-GS, E2VID + 3DGS, and EventNeRF exhibit issues such as blurring, floaters and noise. Notably, the ablation study highlights the impact of removing key components. Removing $L_\text{iso}$ leads to an increase in floaters and artifacts. In contrast, the experiment without adaptive event windows (AW) shows little difference in performance. This is likely due to the relatively consistent camera speeds in this dataset that reduce the potential benefits of AW. 
} 
\label{fig:supp_real_data}
\end{figure*}

\begin{figure*}[h]
\centering
\includegraphics[width=1.0\linewidth]{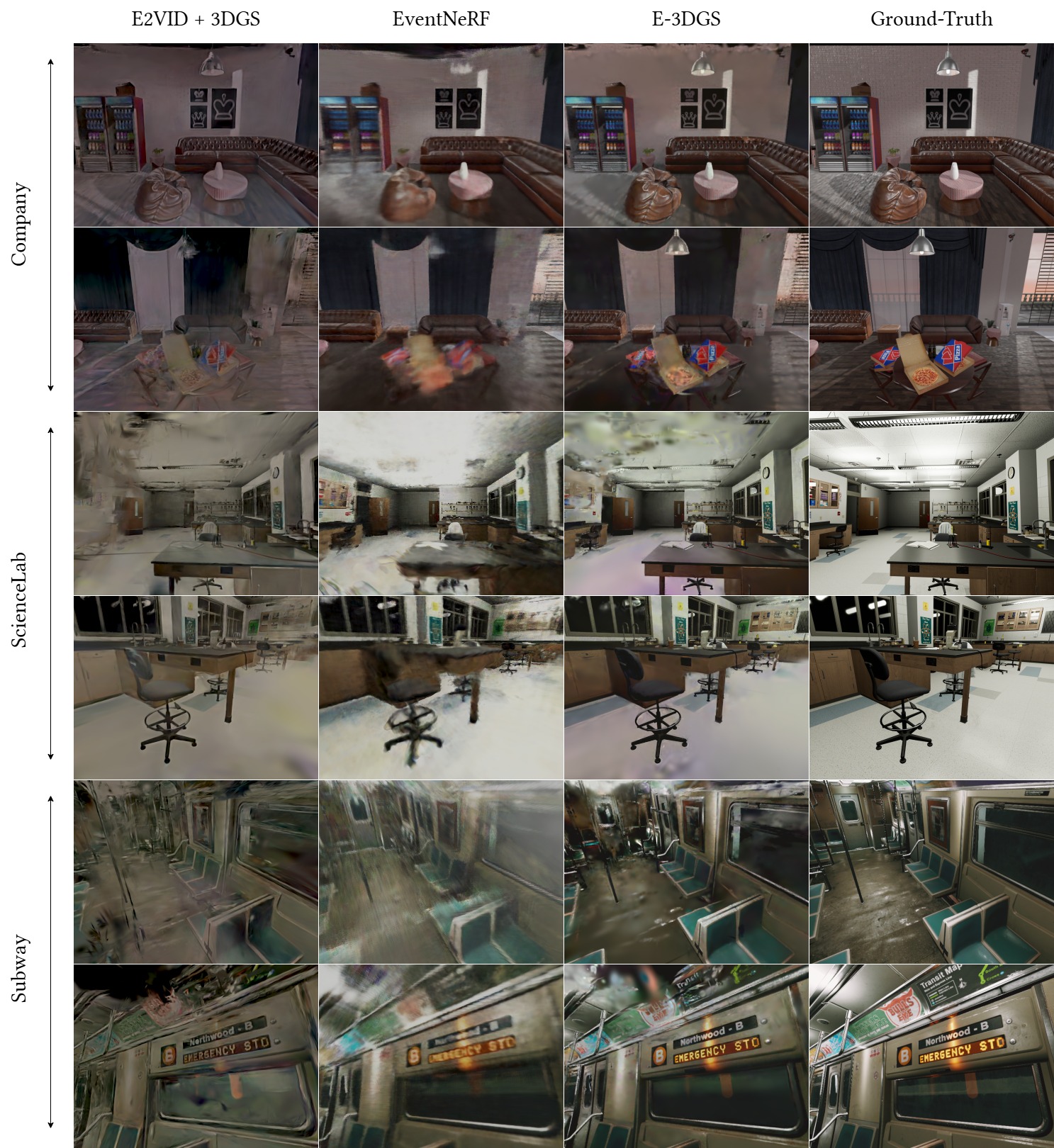}
\caption{Comparison of E-3DGS vs.~baselines on the E-3DGS-Synthetic dataset. As observed in the main paper, E2VID + 3DGS struggles with poor color reconstruction but captures edges and structure reasonably well. EventNeRF suffers from noise and a lack of sharpness. In contrast, our method delivers clear details and accurate colors, with issues mainly confined to less observed areas, such as the roof. Best viewed with zoom.} 
\label{fig:supp_synthetic}
\end{figure*}

\begin{figure*}[h]
\centering
\includegraphics[width=1.0\linewidth]{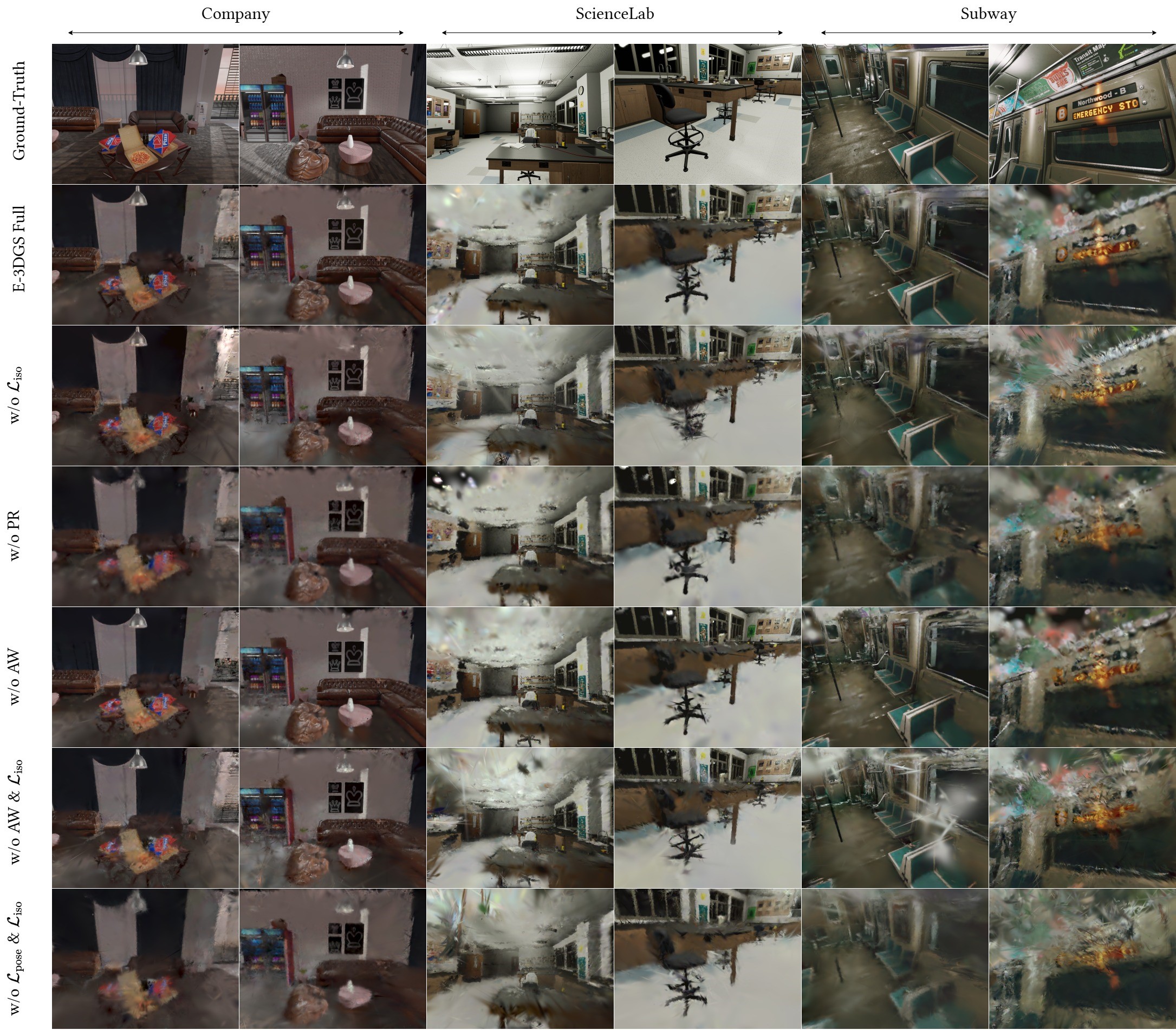}
\caption{
Ablation study of E-3DGS on the E-3DGS-Synthetic-Hard dataset. The increased difficulty of this dataset leads to overall performance deterioration compared to E-3DGS-Synthetic, but our full method still performs well. The version without the adaptive event window (AW) is closest to the full method but shows more artifacts. For example, in the first column of the Company sequence, the sofa shows some artifacts in the AW-removed version that are absent in the full method. Similar minor artifacts are visible elsewhere. The second column of the Subway sequence is interesting, as all versions struggle with reconstructing it. Even so, the full method demonstrates a better structure and fewer artifacts than the others.} 
\label{fig:supp_synthetic_hard}
\end{figure*}

\section{Further Evaluation Details (Ablations)}
\label{sec:supp_evaluation}

To ensure the reliability of the results, all ablation studies are conducted four times, with evaluation metrics averaged to provide more accurate insights and minimize the effects of coincidence. 
For the E-3DGS-Synthetic-Hard dataset, where the camera poses are perturbed, direct evaluation is not feasible due to slight misalignments between the learned 3D scene and the ground truth. 
To correct this, we first freeze the Gaussians and then refine the test poses with a small learning rate to ensure proper convergence. 
This alignment process allows the test views to match the ground truth accurately, enabling precise evaluation. 

\section{Additional Comparisons and Ablations}
\label{sec:additional_comparisons}

In this section, we expand on the main paper experiments by showing additional results on E-3DGS-Real, E-3DGS-Synthetic, and E-3DGS-Synthetic-Hard datasets.
Fig.~\ref{fig:supp_real_data} demonstrates the performance of E-3DGS in comparison to Deblur-GS~\cite{deblurgs}, E2VID~\cite{e2vid}+3DGS~\cite{3dgs} and EventNeRF~\cite{eventnerf} on the E-3DGS-Real dataset.
These baselines exhibit severe artifacts such as blur, floaters and noise.
In the same figure, we also demonstrate the impact of the key components of our method.
Removing $L_\text{iso}$ leads to increased amounts of floaters and other artifacts.
As the captured camera poses contain noise, pose refinement (PR) is crucial to achieve accurate results.
Hence, without it, the model cannot produce accurate predictions, resulting in severe artifacts and blurriness.
However, the model without the adaptive windows (AW) shows similar performance to the full model.
That is likely due to the overall uniformity of the camera speeds in the used dataset, which diminishes the potential impact of adaptive event windows.

In Fig.~\ref{fig:supp_synthetic}, we compare E-3DGS against EventNeRF~\cite{eventnerf} and E2VID~\cite{e2vid}+3DGS~\cite{3dgs} on E-3DGS-Synthetic dataset.
Both baselines perform poorly: While E2VID+3DGS captures the edges and the general structure, it struggles with color representation, and EventNeRF reconstruction is much noisier and blurrier compared to our method.
In contrast, our E-3DGS outperforms them, showing clear and sharp novel views with accurate color representation.
Some issues are still observable but are mostly in less supervised areas, \eg,~on the roof in ScienceLab or Subway scenes.

Lastly, Fig.~\ref{fig:supp_synthetic_hard} visualizes results of the ablation study on the E-3DGS-Synthetic-Hard dataset. 
In comparison to E-3DGS-Synthetic, this dataset has artificially added camera extrinsics noise, which we describe in Sec~\ref{sec:supp_dataset_perturb}, and drastically increased camera speed variation (Sec.~\ref{sec:datasets}). 
While these changes make obtaining high reconstruction quality more difficult, our full method still works well, outperforming all ablated models. 
As on the E-3DGS-Real, removing $L_\text{iso}$ results in severe artifacts (e.g.,~in the first view of Company or in the second view of Subway). 
E-3DGS-Synthetic-Hard dataset has camera pose noise, and, hence, using pose refinement (PR) is important, as removing it results in blurriness and artifacts. 
Removing the adaptive event windows (AW) leads to deterioration; e.g.,~the method without AW exhibits artifacts on the sofa in the first view of the Company sequence that are absent in the results of the full method. 
It is also noteworthy that while all ablated models struggle with the second view of the Subway sequence, the full method, nevertheless,  achieves a better result: The structure is clearer and more recognizable with fewer artifacts.

\end{document}